\documentclass[10pt,journal,compsoc]{IEEEtran}
\ifCLASSOPTIONcompsoc
  \usepackage[nocompress]{cite}
\else
  \usepackage{cite}
\fi
\ifCLASSINFOpdf
\else
\fi
\usepackage{graphicx}
\usepackage{amsmath}
\usepackage{amssymb}
\usepackage{booktabs}
\usepackage{multirow}
\usepackage{makecell}
\usepackage[pagebackref=true,breaklinks=true,letterpaper=true,colorlinks,bookmarks=false]{hyperref}

\def\etal{\emph{et al.}}

\begin{document}
%
\title{Temporally Consistent Video Colorization with Deep Feature Propagation and Self-regularization Learning}
%
%
%
%

\author{
	Yihao Liu$^\star$,
	Hengyuan Zhao$^\star$,
	Kelvin C.K. Chan,
	Xintao Wang,\\
	Chen Change Loy,~\IEEEmembership{Senior Member,~IEEE,}
	Yu Qiao,~\IEEEmembership{Senior Member,~IEEE,}
	Chao Dong
	\thanks{Y. Liu, H. Zhao, C. Dong and Y. Qiao are with Shenzhen Institute of Advanced Technology, Chinese Academy of Sciences, Shenzhen, 518055. E-mail: \{hy.zhao1, chao.dong, yu.qiao\}@siat.ac.cn.}
	\thanks{Y. Liu is also with the University of Chinese Academy of Sciences, Beijing, 100049. E-mail: liuyihao14@mails.ucas.ac.cn.}
	\thanks{Kelvin C.K. Chan and Chen Change Loy are with Nanyang Technological University. E-mail: chan0899@e.ntu.edu.sg, ccloy@ntu.edu.sg. Xintao Wang is with Applied Research Center (ARC), Tencent PCG. E-mail: xintaowang@tencent.com.}
	\thanks{$\star$ Y. Liu and H. Zhao are co-first authors.}}

%
%

\markboth{Journal of \LaTeX\ Class Files,~Vol.~xx, No.~x, August~2021}%
{Liu \MakeLowercase{\textit{et al.}}: Temporally Consistent Video Colorization with Deep Feature Propagation and Self-regularization Learning}
%



\IEEEtitleabstractindextext{%
\begin{abstract}
Video colorization is a challenging and highly ill-posed problem. Although recent years have witnessed remarkable progress in single image colorization, there is relatively less research effort on video colorization and existing methods always suffer from severe flickering artifacts (temporal inconsistency) or unsatisfying colorization performance. We address this problem from a new perspective, by jointly considering colorization and temporal consistency in a unified framework. Specifically, we propose a novel temporally consistent video colorization framework (TCVC). TCVC effectively propagates frame-level deep features in a bidirectional way to enhance the temporal consistency of colorization. Furthermore, TCVC introduces a self-regularization learning (SRL) scheme to minimize the prediction difference obtained with different time steps. SRL does not require any ground-truth color videos for training and can further improve temporal consistency. Experiments demonstrate that our method can not only obtain visually pleasing colorized video, but also achieve clearly better temporal consistency than state-of-the-art methods. Codes will be available. A Video demo is provided  \hyperref{https://www.youtube.com/watch?v=c7dczMs-olE}{category}{name}{here.}
\end{abstract}

}

\maketitle

\IEEEdisplaynontitleabstractindextext

%
\IEEEpeerreviewmaketitle

\IEEEraisesectionheading{\section{Introduction}\label{sec:introduction}}

%
%
%
%
\IEEEPARstart{V}{ideo} colorization aims to generate a fully colored video from its monochrome version. This topic is attractive with wide applications, since there are numerous legacy black-and-white movies produced in the past ages. Colorization can also assist other computer vision tasks such as detection \cite{fasterrcnn,yolo}, tracking \cite{Vondrick_2018_ECCV,siamdw} and video action recognition \cite{Larsson_2017_CVPR}.
\begin{figure}[h]
	\centering
	\includegraphics[width=0.95\linewidth]{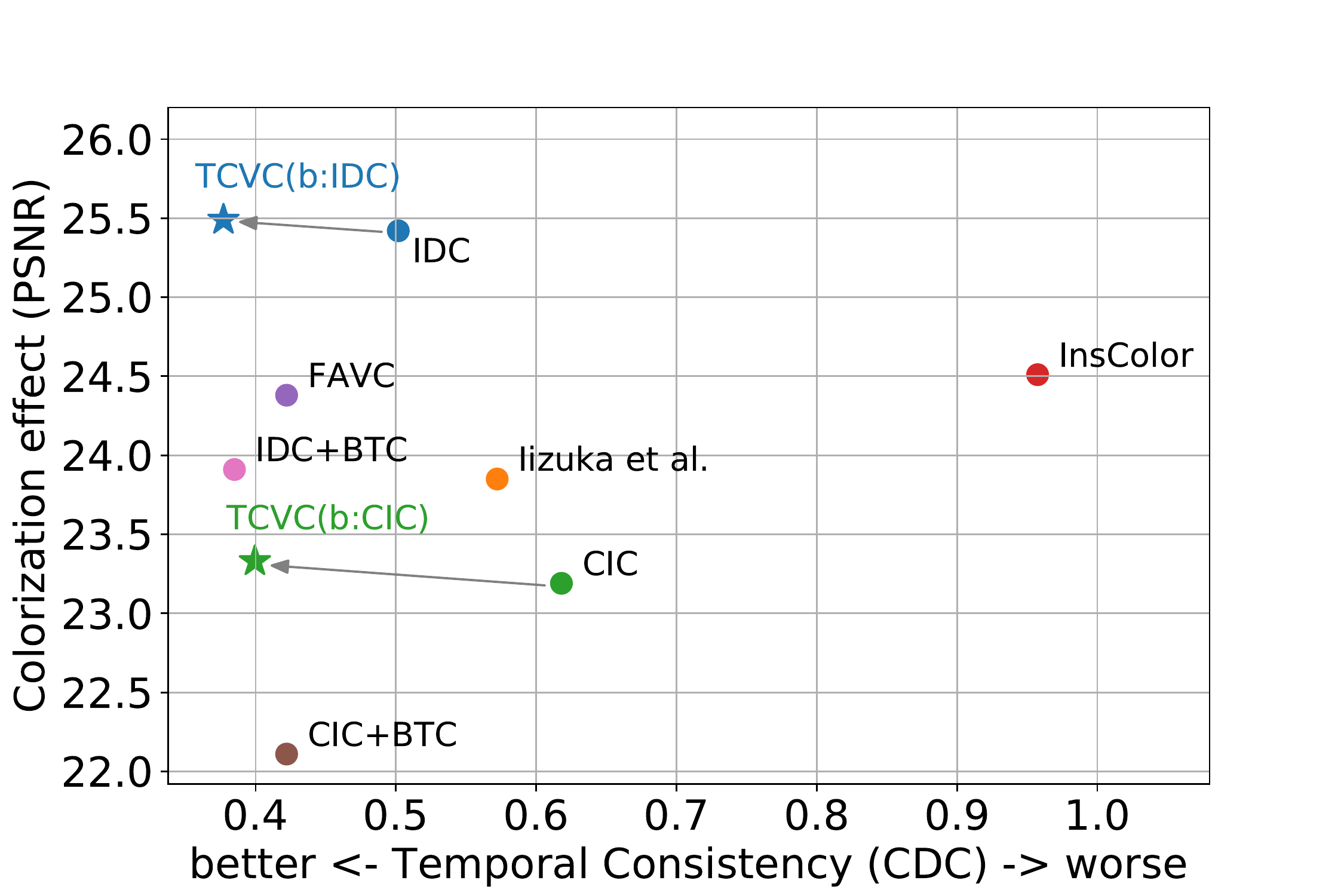}
	\caption{Compared with existing algorithms (CIC \cite{cic}, IDC \cite{tog17}, FAVC \cite{favc} and BTC \cite{btc}), our method achieves both satisfactory colorization performance and good temporal consistency. $b$ denotes the image-based method backbone.}
	\label{fig:chart}
	
\end{figure}

\begin{figure*}[ht]
	\centering
	\includegraphics[width=0.98\linewidth]{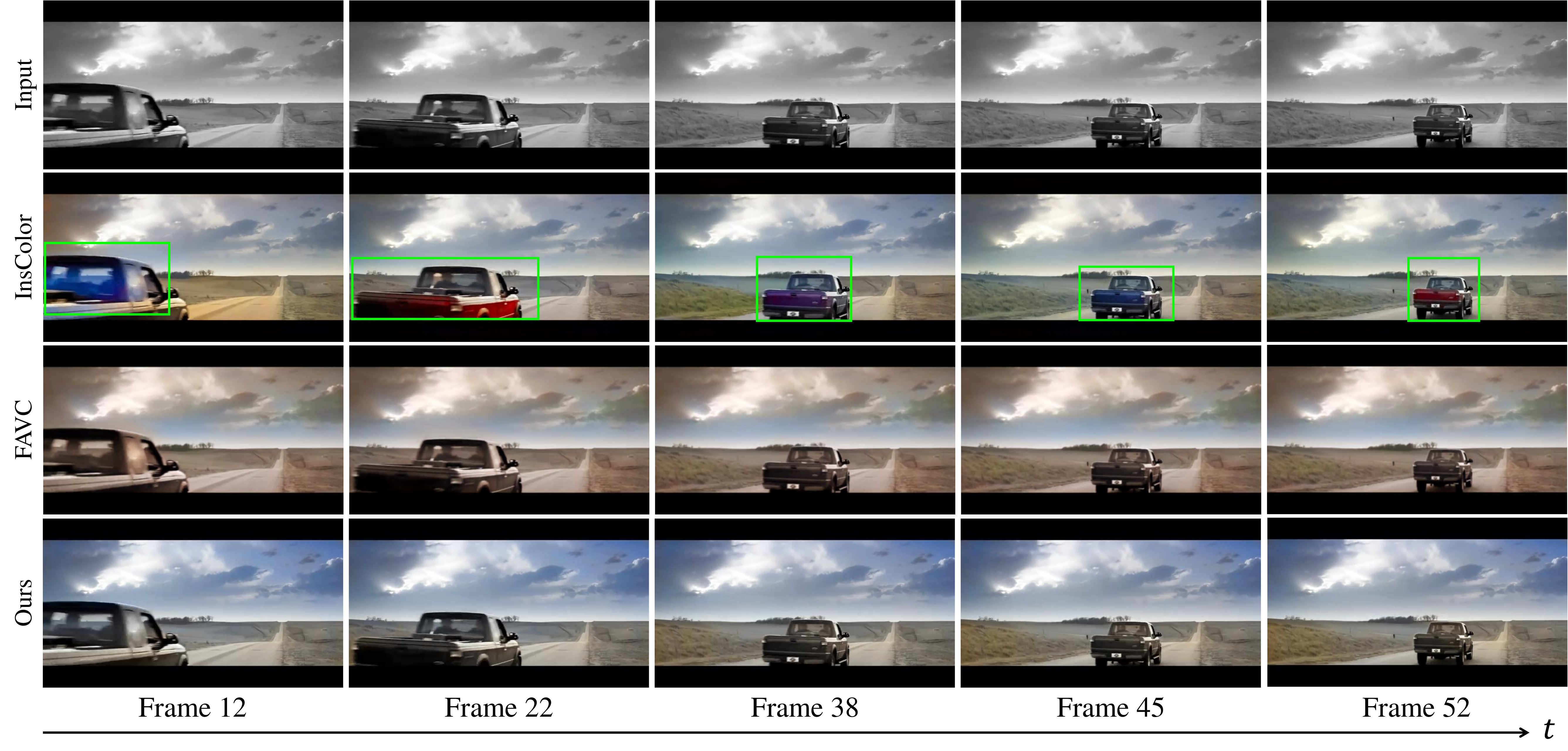}
	\caption{Image-based colorization method, e.g. InsColor \cite{inscolor}, tends to bring about severe flickering artifacts with inconsistent colors (highlighted in green rectangles). The colorization effect of video-based method FAVC \cite{favc} is not satisfactory. The sky is hazy, the grass is not fully colorized and the overall results are grayish. Instead, our method can achieve good temporal consistency while maintaining excellent colorization performance. More comparison results are shown in Section \ref{sec:experiments}.}
	\label{fig:page_one}
	
\end{figure*}

Colorization is a challenging problem due to its highly ill-posed and ambiguous nature. In recent years, plenty of single image colorization methods are proposed and have achieved remarkable progress \cite{let,cic,deepcolor,tog17,inscolor}. Compared with image colorization, video colorization \cite{paul2016spatiotemporal,sheng2013video,favc} is more complex, and receives relatively less attention. It requires not only satisfactory \textbf{colorization performance} but also good \textbf{temporal consistency}, as evaluated in Figure \ref{fig:chart}. A simple way to realize this task is to treat a video sequence as a series of frames and to process each frame independently using an image-based colorization model. In practice, however, when colorizing consecutive sequences, this naive solution tends to produce results suffering from flickering artifacts (temporal inconsistency). As shown in Figure \ref{fig:page_one}, the results of InsColor \cite{inscolor}, a recent state-of-the-art image-based method, are not temporally consistent. Although the colorization effect of each frame is good, the overall results contain unstable flickering, e.g., the colors of the sky and the clothes are inconsistent. This highlights the temporal consistency problem of video colorization.

In general, there are currently two ways to realize temporally consistent video colorization. The first one is to redesign a specialized video colorization model with explicitly considering temporal coherence. It demands tedious domain knowledge to devise the algorithm involving delicate exploration of network structures and loss functions \cite{cic, tog17}. A recent work FAVC \cite{favc} first employs deep learning to achieve automatic video colorization by utilizing self-regularization and diversity loss. However, with their focus mainly on consistency, their colorization performance for individual frame is far from satisfactory. As shown in the thrid row of Figure \ref{fig:page_one}, the sky is grayish and hazy, and the overall results are not vivid. More results of FAVC can be found in Figure \ref{fig:compare}. Its results are usually unsaturated with grayish or yellowish hue. Without good colorization, the temporal consistency will be of less significance.

Another way is to apply post-processing on the output frames and generate a more temporally consistent video \cite{bonneel2015blind,yao2017occlusion,btc,dvp}. For instance, Lai \etal \cite{btc} present a deep network with ConvLSTM module for blind video temporal consistency (BTC), which minimizes the short-term and long-term temporal loss to constrain the temporal stability. Although such post-processing methods can enhance the temporal consistency, they directly regenerate all the frames of the original video, which largely alter the overall frame contents and increase the potential risk of incorrect modification when outliers occur. Moreover, these methods cannot achieve task-specific processing. Because different videos in various tasks are manipulated by the same operators, which could lead to dramatic quantitative performance drop comparing with the original output (see Figure \ref{fig:chart}). Further, methods like BTC \cite{btc} only consider the information of previous frames in forward propagation direction. It is necessary to integrate bidirectional information in handling consecutive video sequence. 

Unlike the aforementioned approaches, we tackle video colorization from a new perspective. Rather than designing a complicated and specialized model, we jointly account for both frame-level colorization and temporally consistent constraints  in a unified deep architecture. Specifically, we propose a novel Temporally Consistent Video Colorization framework (TCVC) that leverages deep features extracted from image-based model $\mathcal{G}$ to generate contiguous adjacent features by bidirectional feature propagation. We only utilize $\mathcal{G}$ to extract several anchor frame features while the remaining internal frame feautres are all generated from anchor frames. Eventually, the colorization performance of our method surpasses that of image-based model $\mathcal{G}$, and the temporal consistency is largely improved as well. Throughout the process, we formulate the spatial-temporal alignment and propagation in high-dimensional feature space rather than image space. Differing from conventional supervised learning, we do not employ any explicit loss with the ground-truth color video, but only adopt the temporal warping loss for self-regularization. As a result, our method is label-free and data-independent. Such self-regularization mechanism also makes the training procedure very efficient. Experiments demonstrate that the proposed framework can favorably preserve the colorization performance of image-based method while simultaneously achieving state-of-the-art temporal consistency for video colorization (as compared in Figure \ref{fig:chart}). 

\section{Related Work}
\textbf{Image and video colorization.} Conventional colorization methods resort to additional information provided by user scribbles \cite{levin2004colorization,qu2006manga,luan2007natural,larsson2016learning,chen2012manifold} or example images \cite{gupta2012image,welsh2002transferring,welsh2002transferring,liu2008intrinsic}. These methods treat colorization as a constrained optimization problem, e.g., Levin \etal \cite{levin2004colorization} proposed an interactive colorization technique that propagated colors from scribbles to neighboring similar pixels. Recently, deep learning techniques have been applied to colorization \cite{deepcolor,let,cic,tog17,he2018deep,lee2020reference,inscolor}. Iizuka \etal \cite{let} devised a two-branch network for jointly learning colorization and classification. Zhang \etal \cite{cic} modeled the colorization as a classification problem to predict the distribution of possible colors for each pixel. Su \etal \cite{inscolor} proposed an instance-aware image colorization model which integrated object detection and colorization together. Aforementioned works have achieved impressive performance on single images but heavily suffer from flickering artifacts when tested on video. Video colorization \cite{sheng2013video,paul2016spatiotemporal,xu2020stylization,deepexemplar,favc} needs to consider both colorization performance and temporal consistency. Recently, a pioneer deep-learning-based work FAVC \cite{favc} was proposed for automatic video colorization, which is the most relevant work to ours. FAVC regularized its model with KNN graph built on the ground-truth color video and simultaneously posed a temporal loss term for constraining temporal consistency. However, the colorization performance of FAVC is not satisfactory.

\textbf{Video temporal consistency.} The temporal consistency problem is addressed on a diverse type of applications, such as artistic style transfer \cite{gatys2015neural,zhu2017unpaired,artvideo,vpn,chu2020learning}, image enhancement \cite{fdgan,csrnet,eilertsen2019single,lei2020blind} and colorization \cite{btc,favc}. Bonneel \etal \cite{bonneel2015blind} proposed a gradient-domain technique to infer the temporal regularity from the original unprocessed video. Yao \etal \cite{yao2017occlusion} developed an online keyframe strategy to keep track of the dynamic objects and to handle occlusions. Lai \etal \cite{btc} presented a ConvLSTM-based method which took advantage of deep recurrent network and perceptual similarity \cite{johnson2016perceptual}. These video temporal consistency algorithms are usually post-processing methods, which modify each frame of the input video and produce a new output video. Instead of applying post-processing, our work addresses the temporal consistency and video colorization in a unified framework.

\section{Methodology}
\begin{figure*}[t]
	\centering
	\includegraphics[width=0.86\linewidth, height=0.35\textheight]{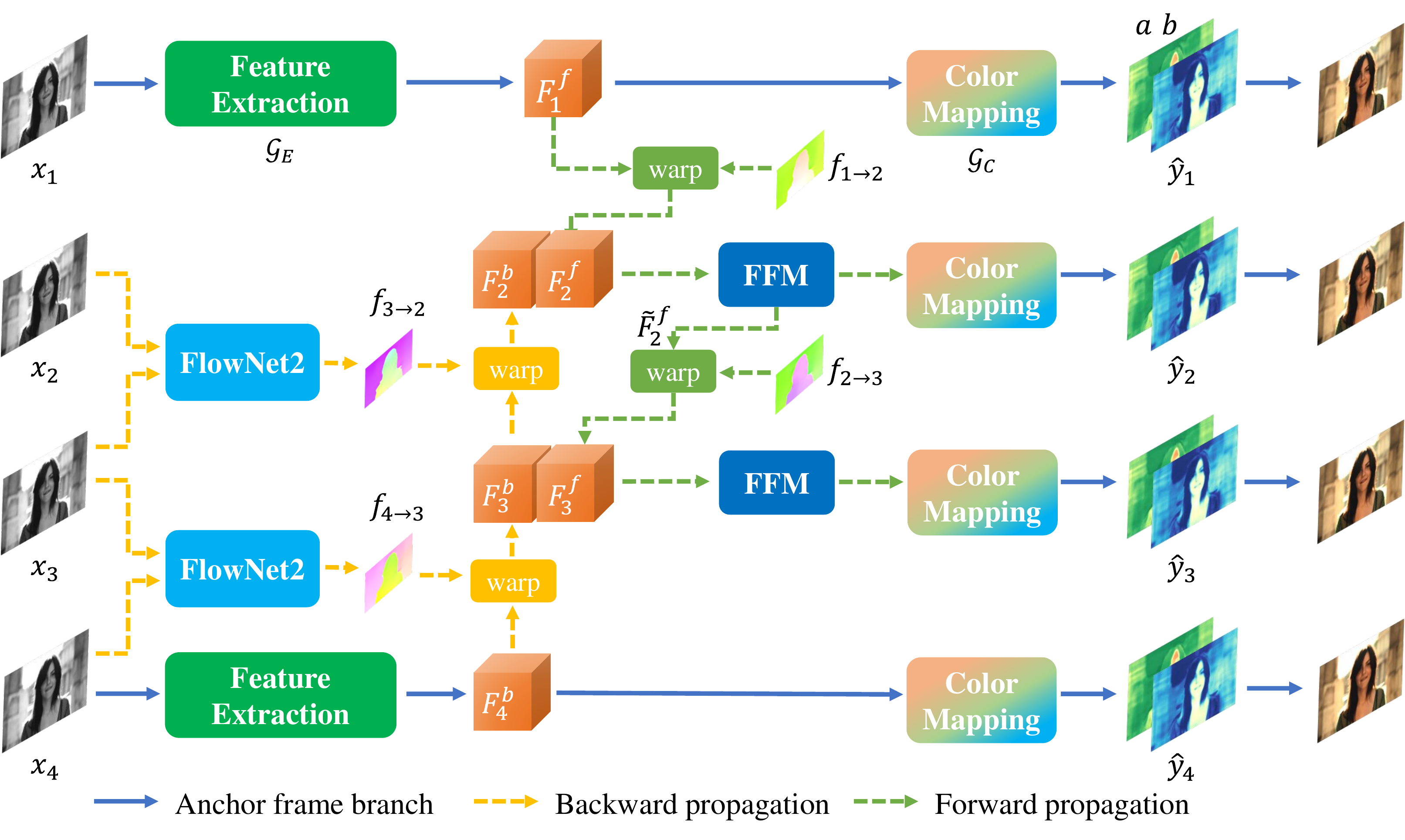}
	\caption{The proposed TCVC framework (take $N=4$ for example). The anchor frame branch colorizes the two anchor frames and extracts the deep features for propagation. With bidirectional deep feature propagation, the internal frame features are all generated from anchor frames, which ensures the temporal consistency in high-dimensional feature space.
	}
	\label{fig:framework}
	
\end{figure*}

\subsection{Temporally Consistent Video Colorization}
Given an input grayscale video, the objective of video colorization is to obtain its corresponding colorized version. Following previous works \cite{let,cic,tog17,inscolor}, we perform this task in CIE $Lab$ color space and predict two associated chrominance channels of a grayscale image, i.e., $a$ channel and $b$ channel.
\subsubsection{Overview}
For a long input grayscale video sequence, we can decompose it into several intervals. Assume each interval sequence contains $N$ consecutive grayscale frames ${\textbf{X}=\{x_1, x_2, \cdots, x_N\}}$, we denote the start frame $x_1$ and the last frame $x_N$ as \textit{anchor frames}, and the remaining $N-2$ frames as \textit{internal frames}. Thereupon, the input sequence is divided with several anchor frames and internal frames in between. Our method works on each interval sequence and we consider the input sequence as a continuum with continual camera and object motions.

The proposed TCVC framework leverages the temporal and motional information of consecutive frames in high-dimensional feature space. For any image-based colorization model $\mathcal{G}$, it can be naturally separated into two parts: feature extraction module $\mathcal{G}_E$ and color mapping module $\mathcal{G}_C$. Generally, the color mapping module corresponds to the last output layer of $\mathcal{G}$, while the feature extraction module includes all the layers before the output layer. As illustrated in Figure \ref{fig:framework}, firstly, the deep features of the two anchor frames are extracted through the feature extraction module $\mathcal{G}_E$. Then, the features are sequentially propagated in forward and backward directions frame by frame. They contain essential information for colorization. Finally, at each frame step, the associated deep features are fed into the shared color mapping module $\mathcal{G}_C$ to obtain the predicted color chrominance channels $y_i$.

\subsubsection{Anchor Frame Processing}
Given the anchor frames at both ends of the interval sequence, our goal is to generate the color channels of each internal frame by propagating the features of the anchor frames. The anchor frames are directly processed by colorization backbone $\mathcal{G}$. As shown in Figure \ref{fig:framework}, in each interval sequence, the anchor frame branch colorizes the two anchor frames and extracts the deep features for propagation:
\begin{equation} \label{equ:anchor}	
\begin{array}{lcl}
F_1^f = \mathcal{G}_E(x_1),  \quad \hat{y}_1 = \mathcal{G}_C(F_1^f),\\
F_N^b = \mathcal{G}_E(x_N),  \quad \hat{y}_N = \mathcal{G}_C(F_N^b).
\end{array}
\end{equation}
The superscripts $f$ and $b$ represent forward and backward directions, respectively. As described in Equation (\ref{equ:anchor}), the extracted features here are the output of the penultimate layer of model $\mathcal{G}$, which properly matches with the color mapping module $\mathcal{G}_C$. The color mapping module $\mathcal{G}_C$ accepts these features and outputs the predicted color channels. Remarkably, $\mathcal{G}$ can be any CNN-based algorithm, such as CIC \cite{cic}, IDC \cite{tog17}, etc., making it a plug-and-play framework. Note that, in TCVC framework, the colorization backbone $\mathcal{G}$ is fixed without training. We only adopt its $\mathcal{G}_E$ to extract anchor frame features and $\mathcal{G}_C$ to predict $y_i$. Since the anchor frame branch applies model $\mathcal{G}$ on the anchor frames straightforward, it will not change any of the colorization style and performance of $\mathcal{G}$. With the initial deep features extracted from anchor frames, the features of each internal frame are progressively generated by bidirectional propagation. The forward feature propagation is initiated at the start anchor frame $x_1$ and the backward feature propagation is initiated at the last anchor frame $x_N$.

\subsubsection{Bidirectional Deep Feature Propagation} \label{sec:dfp}
For internal frames, we make use of the temporal and motional characteristics of video sequence to generate the associated features from anchor frame features. This procedure is carried out by backward propagation and forward propagation sequentially. We adopt bidirectional feature propagation since all the information needed to generate the internal frame features is encoded and contained in the forward and backward directions. The feature propagation initiates from the backward direction.

\textbf{Backward propagation.} As depicted in Figure \ref{fig:framework}, the feature propagation begins at backward direction. We first estimate the optical flow between two adjacent frames. Based on the estimated motion fields, we can obtain coarsely warped internal frame features in backward direction:
\begin{equation}	
F_i^b = warp(F_{i+1}^b, f_{i+1 \rightarrow i}),
\end{equation}
where $F_i^b$ is the backward warped feature at the $i$-th frame, $i=N-1, N-2, \cdots, 2$. $warp(\cdot, \cdot)$ denotes the warping function, which can be implemented using bilinear interpolation \cite{jaderberg2015spatial} and $f_{i+1 \rightarrow i}$ represents the optical flow from frame $x_{i+1}$ to $x_i$. We adopt FlowNet2 \cite{ilg2017flownet} to compute the flow, due to its effectiveness in relative tasks. The backward propagation starts from $F_N^b $ and generates the features of each internal frame ${F_i^b}$. However, only backward propagation is insufficient. Without complementary information from the opposite direction, the warping operation will cause the features to continuously shift in one direction, resulting in information loss. Therefore, after the backward propagation, the forward propagation will start from the other direction.

\textbf{Forward propagation.} The forward propagation starts from the first frame $x_1$. Similar to the backward propagation, the forward propagation first obtains a coarsely warped internal frame feature based on the estimated optical flow. Furthermore, forward propagation is obligated for more functions, including integrating backward and forward features, and generating color channels with the fused features. To integrate the features propagated in backward and forward directions, we devise an effective frame-specific feature fusion module (FFM). The details of FFM will be described later. We denote the output feature of FFM in the forward propagation as $\tilde{F_i^f}$, which combines fine bidirectional information for subsequent colorization. Note that, except the first forward feature $F_1^f$ is directly delivered to the next frame, the other forward features to be propagated are the features after fusion, i.e., $\tilde{F_i^f}$:
\begin{equation}
F_{i+1}^f = warp(\tilde{F_i^f}, f_{i \rightarrow i+1}),
\end{equation}
where $i=1,2,\cdots,N-2$. After feeding $\tilde{F_i^f}$ into the shared color mapping module $\mathcal{G}_C$, the predicted color channels of internal grayscale frame $x_i$ is obtained:
\begin{equation}
\hat{y}_i = \mathcal{G}_C(\tilde{F_i^f}),
\end{equation}
where $i=2,3,\cdots,N-1$. 

\begin{figure}[htbp]
	\centering
	\includegraphics[width=8.3cm]{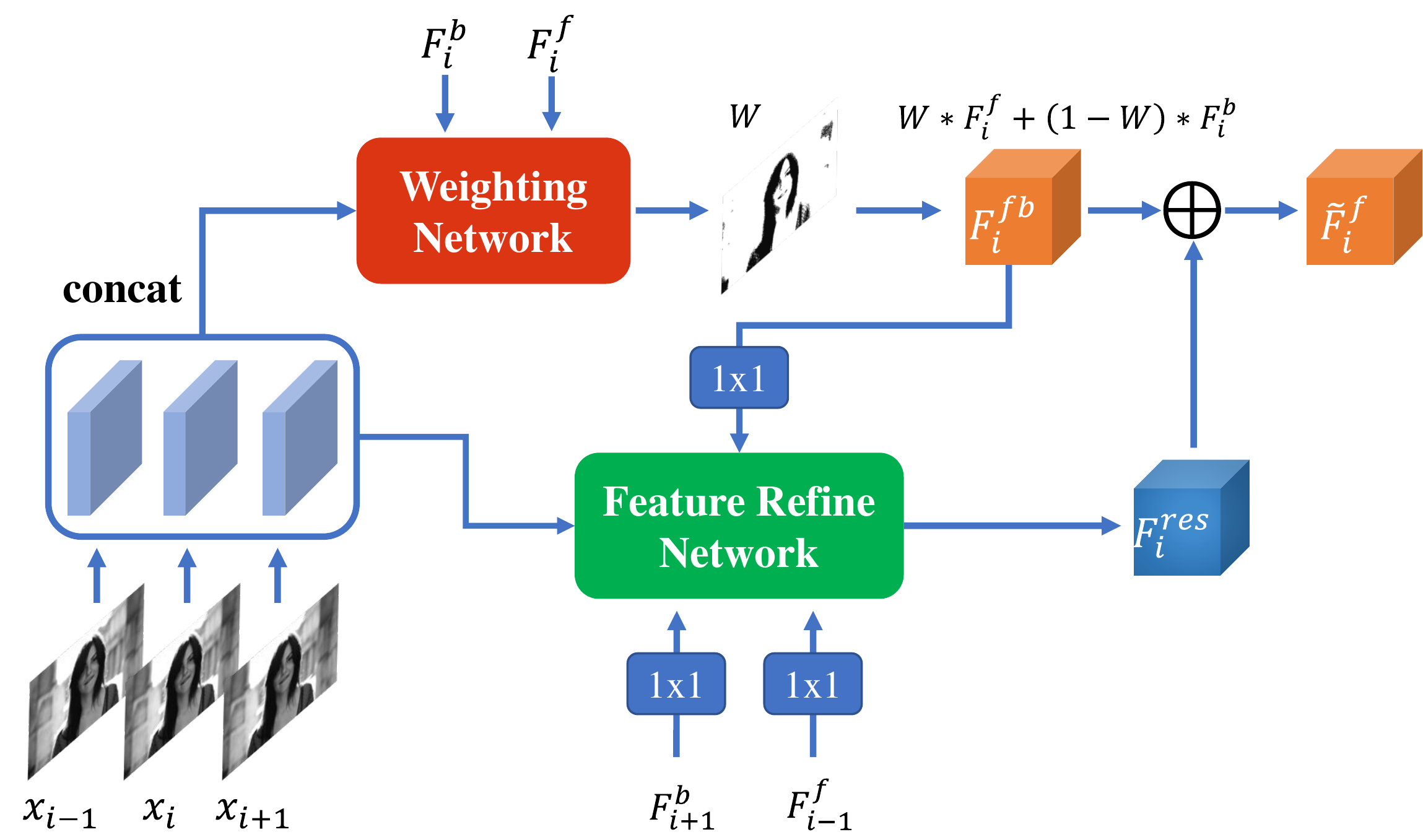}
	\caption{The structure of feature fusion module (FFM), which contains a weighting network and a feature refine network.}
	\label{fig:FFM}
\end{figure}

\textbf{Feature fusion module.} The structure of the proposed feature fusion module is detailed in Figure \ref{fig:FFM}. It contains a weighting network (WN) and a feature refine network (FRN), which are both \textit{three-layer plain CNNs}. In FFM, three consecutive images $x_{i-1}$, $x_{i}$, $x_{i_1}$ are first fed into $\mathcal{G}_E$ to obtain the corresponding features, which are then concatenated together with other inputs to feed in the WN and FRN.

Intuitively, the warped backward feature $F_i^b$ and forward feature $F_i^f$ are both coarsely aligned with current frame $x_i$. However, due to different propagation directions, there are complementary and redundant parts between them in different pixel locations. The weighting network predicts a weighting map $W \in \mathbb{R}^{H \times W \times 1}$ ranged in $[0,1]$. Then, the forward and backward features are fused by a simple linear interpolation: $F_{i}^{fb} = W \odot F_i^f + (1-W) \odot F_i^b,$ where $\odot$ denotes the element-wise multiplication operation. $F_{i}^{fb}$ contains the information of both forward and backward features. Due to the inaccurate flow estimation and the information loss caused by warping operation, errors will accumulate in the propagation process. Therefore, we further refine the feature according to the adjacent spatiotemporal information. 

As shown in Figure \ref{fig:FFM}, the feature refine network accepts the roughly fused feature $F_{i}^{fb}$ and generates a refining residual $F_{i}^{res}$. Specially, the feature refine network additionally takes into account the backward feature $F_{i+1}^{b}$ of the latter frame and the forward feature $F_{i-1}^{f}$ of the previous frame. The reason for such design is that $F_{i+1}^{b}$ and $F_{i-1}^{f}$ implicitly encode and contain all the information needed to obtain the aligned feature at current $i$-th frame. $1 \times 1$ convolutions are used to reduce and unify the dimensionality. The final refined feature at $i$-th frame is obtained as:
\begin{equation}
\tilde{F_i^f} = F_{i}^{fb} + F_{i}^{res}.
\end{equation}
$\tilde{F_i^f}$ will be propagated to the next frame. By utilizing the information of current frame and adjacent frames, FFM can achieve frame-specific feature fusion in a coarse-to-fine manner. With bidirectional feature propagation, the internal frame features are all generated from anchor frame features.

\subsection{Self-regularization Learning}\label{sec:self-reg}
One unique characteristic of the proposed framework is that it utilizes a self-regularization learning scheme without relying on ground-truth color videos. Here the self-regularization learning means that we do not employ any explicit loss with the ground-truth color video, which is different from the conventional supervised learning. In TCVC, we do not need to train the colorization backbone $\mathcal{G}$. To let the network learn temporal consistency, we adopt the temporal warping loss as follows:
\begin{equation}
L_{TW} = \sum_{d=\{1,2\}} \sum_{i=1}^{N-d}\left\| M_{i+d \rightarrow i}\odot (\hat{y}_i-\hat{y}_{i+d}^{warp})  \right\|_2,
\end{equation}
where $\hat{y}_{i+d}^{warp} = warp(\hat{y}_{i+d}, f_{i+d\rightarrow i})$, and $d$ represents the time interval for temporal warping. $M_{i+d \rightarrow i}=\exp(-\alpha \| {y}_i-\hat{y}_{i+d}^{warp} \|_2^2)$ is the visibility mask. Following \cite{btc}, we set $\alpha=50$. This loss function explicitly poses a penalty to the temporal consistency between adjacent frames. It is noteworthy that there is no ground-truth color video used during training. A consecutive grayscale input video is all we need. More discussions can be found in the supplementary file.

Further, the self-regularization learning also makes our framework free from the influence of training and testing data, i.e., the proposed method is data-independent and label-free. As long as the input video contains consecutive motional frames, it can be adopted as our training set. Another advantage of the proposed self-regularization is that it does not require magnanimous training data and it has few trainable parameters. Thus, the training procedure can be very efficient (about two days).

\subsection{Multiple Anchor Frame Sampling}
For a long input sequence (over dozens of frames), we first divide it into several intervals by uniformly sampling anchor frames or specifying the interval length $N$. Once the interval length $N$ is determined, the anchor frames are also determined during inference phase. Empirically, this scheme has already worked fine. However, we also provide an optional post-processing scheme to further enhance the performance. Specifically, we sample the anchor frames multiple times (choose different $N$), and then average the results of each output. This procedure can be regarded as an ensemble method during testing. It could eliminate the uncertainty and inconsistency of anchor frames to some extent and achieve better temporal consistency. In the experiments, we adopt $N=15$ and $N=17$ for ensembling.

\subsection{Discussions}
\subsubsection{Differences with other methods}
The proposed TCVC framework is conceptually different from previous works on video colorization and video temporal consistency in motivation and methodology. We address the video colorization problem from a new perspective. In summary, the proposed method is different from previous solutions in three aspects: i) The proposed framework takes advantage of an ingenious image-based model and focuses on the temporally consistent constraints. It can favorably achieve both good colorization performance and satisfactory temporal consistency. ii) We formulate the spatial-temporal alignment and propagation in high-dimensional feature space. iii) Different from conventional supervised learning, it adopts a self-regularization learning scheme without relying on ground-truth color videos.
\subsubsection{Uniqueness of TCVC}
Our goal is to improve the temporal consistency based on image-based model $\mathcal{G}$ without retraining it. The unique feature of TCVC is that it requires no GT color videos during training. This unique trait is a concomitant by-product, introduced by the method itself with feature propagation and self-regularization learning. Specifically, TCVC first leverages a pretrained image-based model $\mathcal{G}$ to extract anchor frame features containing color information. Then, TCVC propagates the color information from anchor frames to the remaining internal frames. The color information is inherited from $\mathcal{G}$, and TCVC focuses on the propagation of such sparse information with explicitly considering the temporal consistency. The temporal warping loss poses a penalty to implicitly guide the network to better propagate the color information in a self-regularization and self-supervised manner, with explicitly enhancing the temporal consistency between adjacent frames. Further, we have also conducted experiments to demonstrate the effectiveness of such learning scheme.

With such a unique characteristic, TCVC obtains state-of-the-art results on temporally consistant video colorization. As shown in Figure \ref{fig:page_one}, existing image-based method like InsColor \cite{inscolor} tends to produce severe flickering artifacts with inconsistent colorization. The colorization performance of video-based method FAVC \cite{favc} is not satisfactory, which will produce grayish and unsaturated hue. Instead, the colorization effect of TCVC framework is temporally consistent and visually pleasing, due to bidirectional feature propagation and self-regularization learning.

\begin{figure*}[htbp]
	\centering
	\includegraphics[width=0.98\linewidth,height=4in]{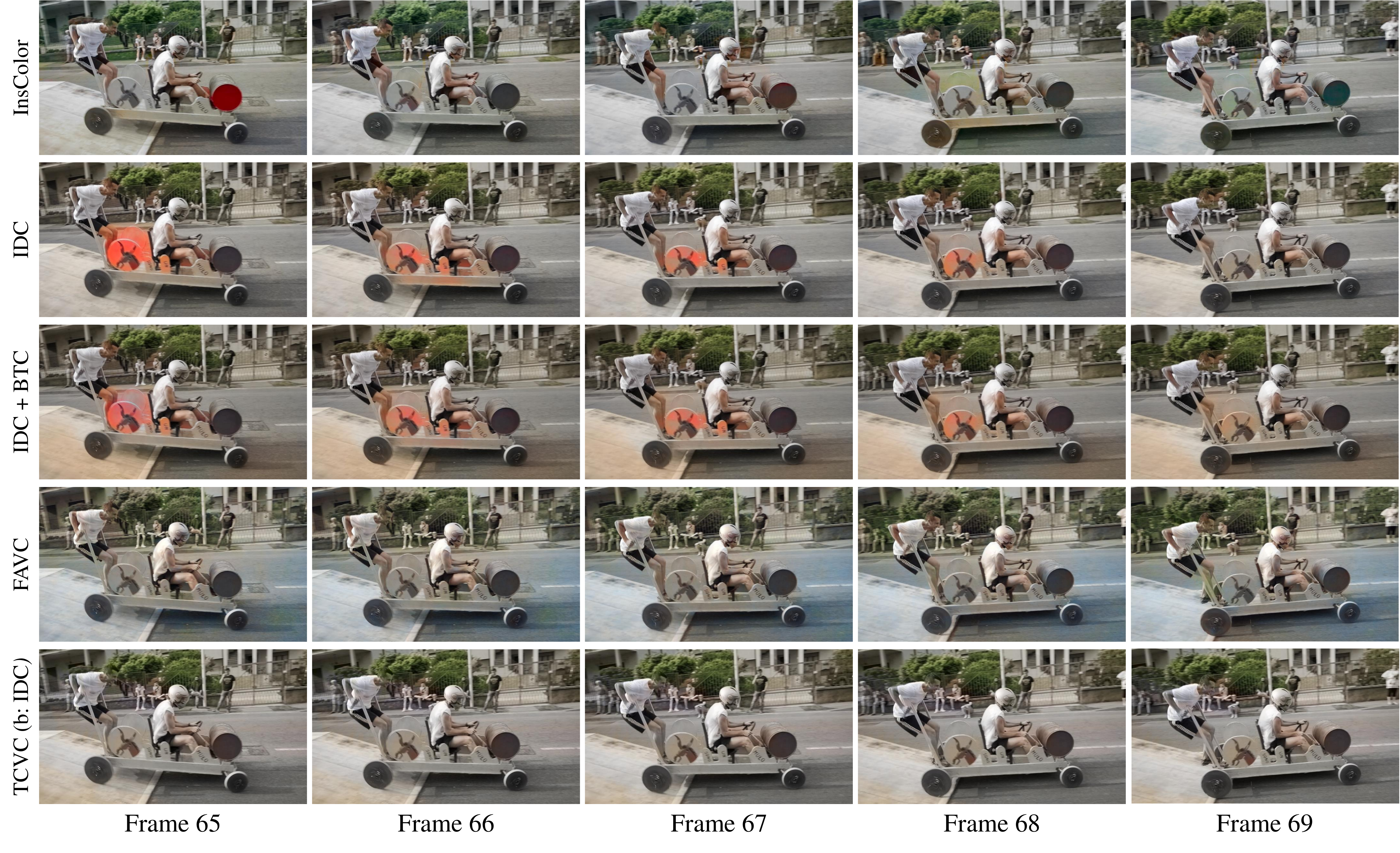}
	\includegraphics[width=0.98\linewidth,height=4in]{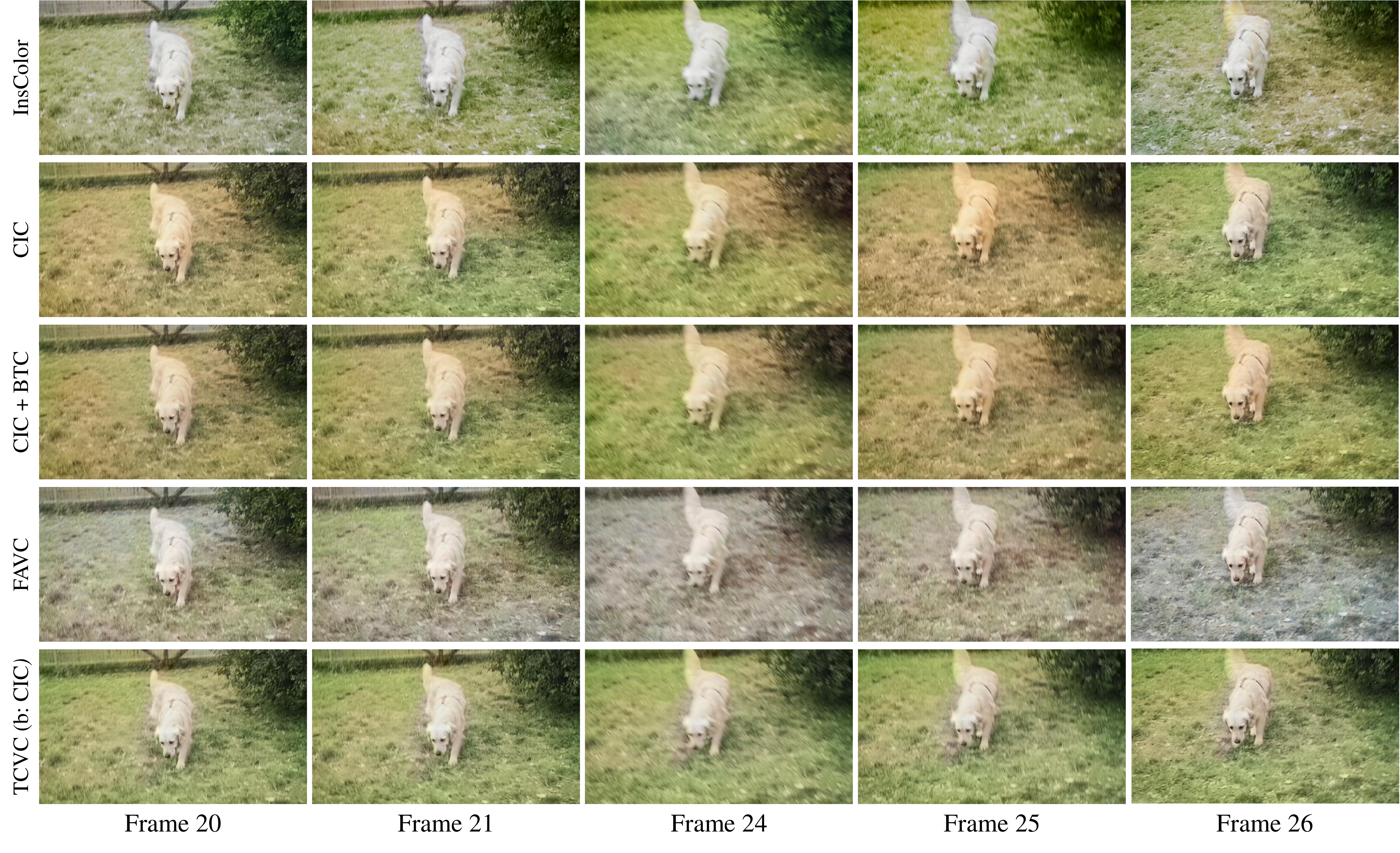}
	\caption{Visual comparison with state-of-the-art methods. Image-based methods \cite{inscolor,tog17} are prone to produce severe flickering artifacts. Post-processing method BTC \cite{btc} cannot achieve long-term temporal consistency well and cannot handle outliers. The results of FAVC \cite{favc} are usually unsaturated and sometimes contain strange greenish hue, e.g., there are strange greenish regions on the gun in the upper sequence. Please zoom in for best view.}
	\label{fig:compare}
	
\end{figure*}

\begin{figure*}[htbp]
	\centering
	\includegraphics[width=0.98\linewidth,height=3.65in]{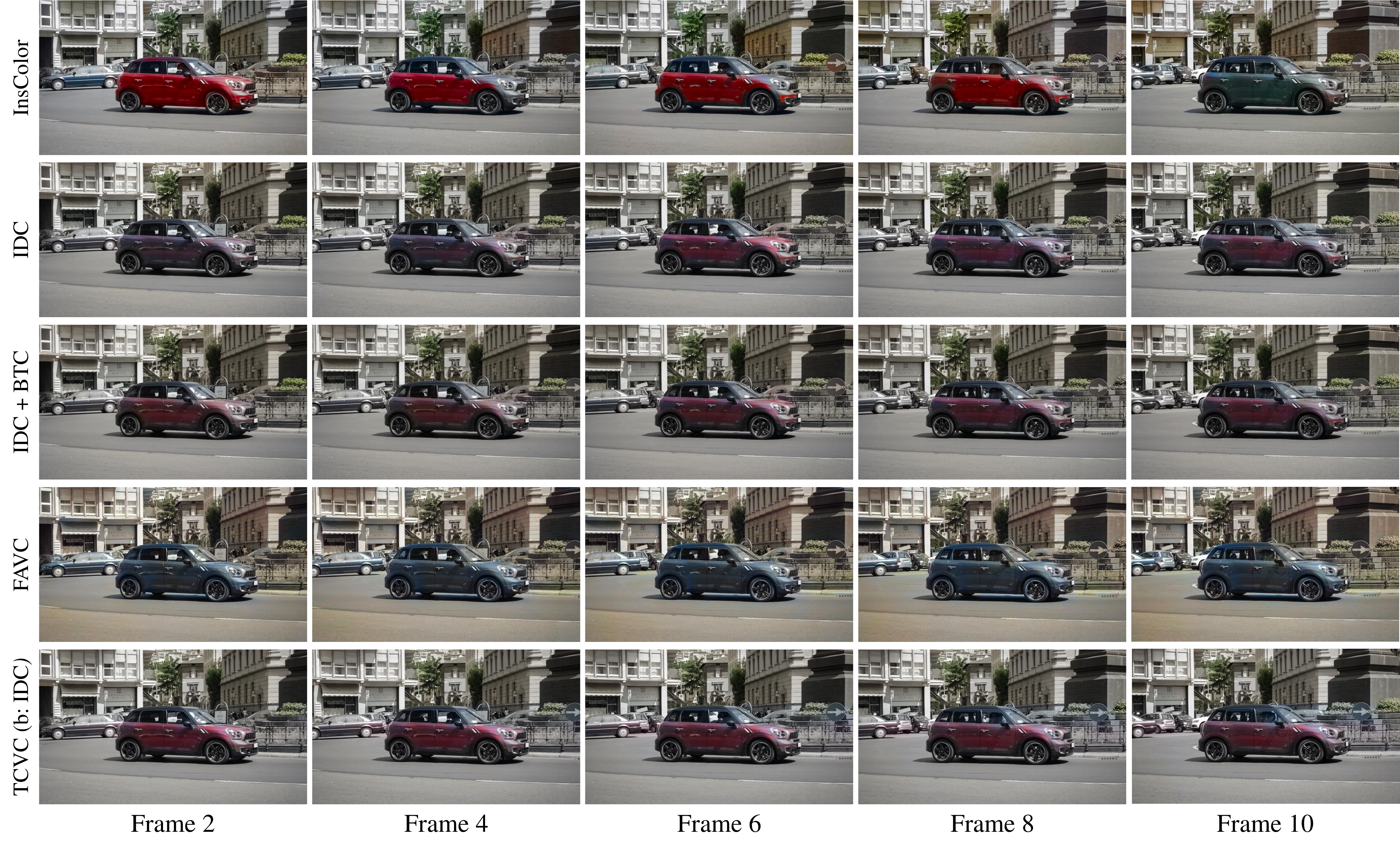}
	\vspace{5pt}
	
	\includegraphics[width=0.98\linewidth,height=3.65in]{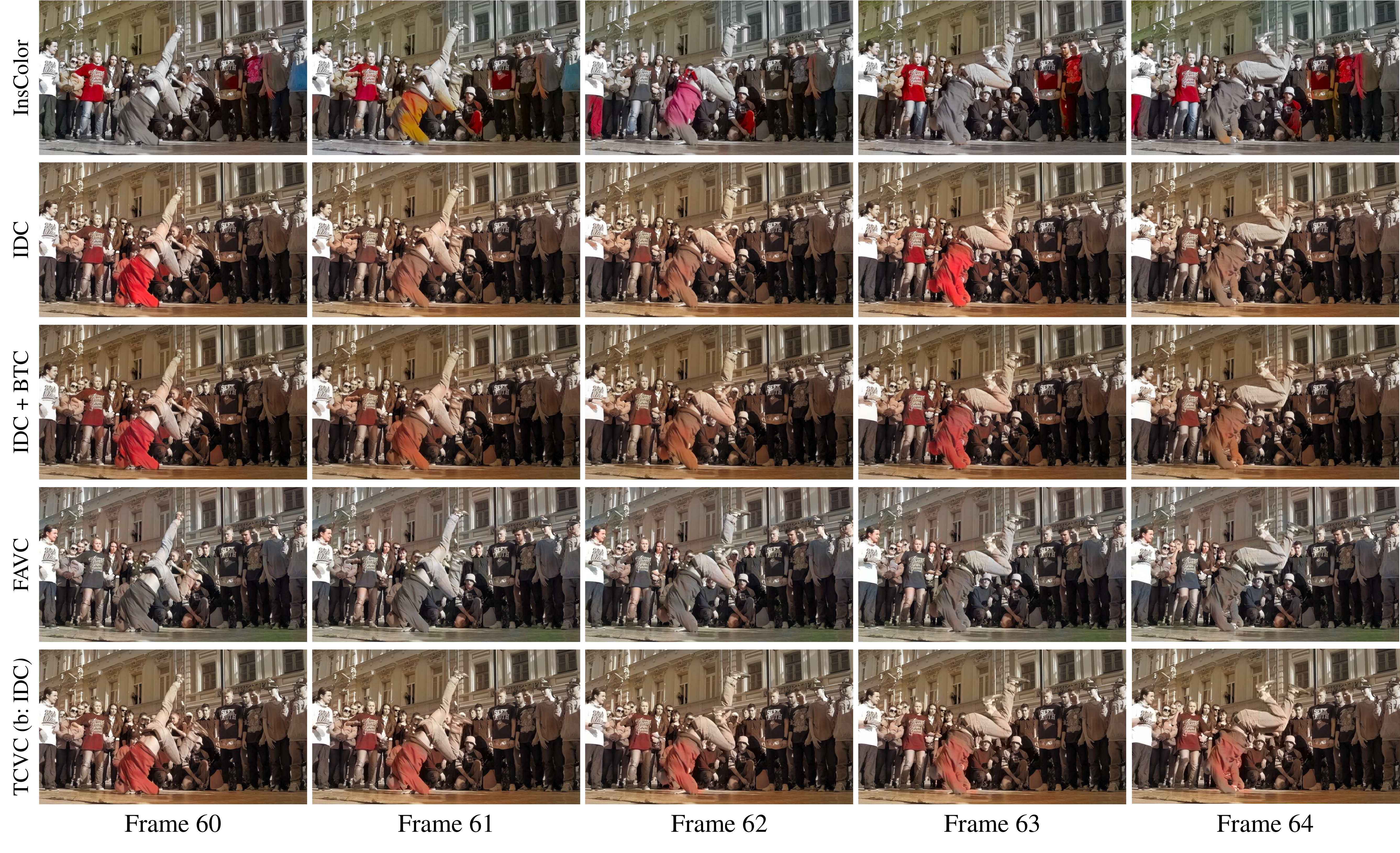}
	\caption{Visual comparison with state-of-the-art methods.}
	\label{fig:compare1}
	
\end{figure*}

\setlength{\tabcolsep}{2.8pt}
\begin{table*}[]
	\begin{center}
		\begin{tabular}{@{}lccccccccccc@{}}
			\toprule
			\multirow{2}{*}{Method} & \multicolumn{5}{c}{DAVIS30 (medium frame length)}                & \multicolumn{1}{l}{} & \multicolumn{5}{c}{Videvo20 (long frame length)}               \\  
			& \underline{Warp Error}$\downarrow$ & \multicolumn{1}{c}{\underline{CDC}$\downarrow$}      & \underline{PSNR}$\uparrow$  & \underline{L2 Error}$\downarrow$ & \underline{Colorfulness}$\uparrow$ & \multicolumn{1}{l}{} & \underline{Warp Error}$\downarrow$ & \multicolumn{1}{c}{\underline{CDC}$\downarrow$}   & \underline{PSNR}$\uparrow$  & \underline{L2 Error}$\downarrow$ & \underline{Colorfulness}$\uparrow$ \\ \midrule
			Iizuka \etal \cite{let}                 & 0.001467    & 0.005722 & 23.85 & 12.77 & 20.16    &                      & 0.001127    & 0.004821 & 23.69 & 12.94 & 20.84    \\
			CIC \cite{cic}                 & 0.001431    & 0.006180 & 23.19 & 15.88  & 30.34   &                      & 0.001081    & 0.003595 & 22.51 & 17.11  & 29.19  \\
			IDC \cite{tog17}                   & 0.001271    & 0.005017 & 25.42 & 11.99   & 21.70  &                      & 0.000977    & 0.002578 & 25.35 & 11.69  & 19.07   \\
			InsColor \cite{inscolor}                  & 0.001957    & 0.009574 & 24.51 & 13.20  & 19.54   &                      & 0.001599   & 0.008019 & 24.80 & 12.16  & 20.37   \\ \midrule
			\cite{cic} + BTC \cite{btc}             & 0.001019    & 0.004201 & 22.11 (-1.08) & 16.67 & 30.00 (-0.34)    &                      & 0.000720    & 0.001874 & 21.72 (-0.79) & 17.68   & 27.81 (-1.38)  \\
			\cite{tog17} + BTC \cite{btc}      & \textbf{0.000964}    & 0.003849 & 23.91 (-1.51) & 12.91  & 20.65 (-1.05) &                      & \textbf{0.000694}    & 0.001685 & 23.81 (-1.54) & 12.60  & 17.90 (-1.17)  \\ 
			\cite{cic} + DVP \cite{dvp}             & 0.001505    & \textbf{0.003447} & 23.07 (-0.12) & 15.69 & 27.61 (-2.73)   &                      & 0.000968    & 0.001859 & 21.71 (-0.80) & 18.89  & 31.30 (+2.11)  \\
			\cite{tog17} + DVP \cite{dvp}             & 0.001538    & 0.003539 & 24.00 (-1.42) & 14.28  & 21.70 (-0.00)  &                      & 0.000985    & 0.001708 & 25.23 (-0.12) & 11.71  & 18.50 (-0.57)   \\ \midrule
			FAVC \cite{favc}                    & 0.001332    & 0.004221 & 24.38 & 13.26    & 18.55 &                      & 0.001052    & 0.001880 & 24.81 & 12.21  & 16.28   \\ \midrule
			TCVC ($b$: \cite{cic})            & 0.001145    & 0.003995 & 23.33 (+0.14) & 15.56  & 30.34 (-0.00)  &                      & 0.000902    & 0.001902 & 22.71 (+0.20) & 16.74  & 29.19 (-0.00)  \\
			TCVC ($b$: \cite{tog17})             & 0.001139    & 0.003771 & 25.49 (+0.07) & 11.90 & {21.70 (-0.00)}   &                      & 0.000896    & 0.001691 & 25.39 (+0.04)& 11.66  & 19.07 (-0.00)   \\
			TCVC$^+$ ($b$: \cite{cic})     & 0.001136    & 0.003940 & 23.44 (+0.25) & 15.41   & {30.34 (-0.00)} &                      & 0.000899    & 0.001864 & 22.71 (+0.20)& 16.76 & {29.19 (-0.00)}  \\
			TCVC$^+$ ($b$: \cite{tog17})   & 0.001135    & {0.003733} & \textbf{25.50 (+0.08)} & \textbf{11.86} & {21.70 (-0.00)}   &                      & 0.000894    & \textbf{0.001649} & \textbf{25.43 (+0.08)} & \textbf{11.59}  & {19.07 (-0.00)}  \\ \bottomrule
		\end{tabular}
	\end{center}
	\caption{Quantitative performance on DAVIS30 and Videvo20 datasets. Applying BTC \cite{btc} improves the temporal consistency but decreases the PSNR values dramatically. TCVC framework can favorably achieve both satisfactory colorization effect and temporal consistency. $b$ indicates the backbone we choose for TCVC and $^+$ denotes adopting multiple anchor frame sampling ensemble.}\label{tab:quantitative_compare}
\end{table*}

\section{Experiments}\label{sec:experiments}
\textbf{Datasets.} Following previous works \cite{btc,favc}, we adopt DAVIS dataset \cite{DAVIS} and Videvo dataset \cite{btc} for training and testing. The DAVIS dataset is designed for video segmentation, which includes a variety of moving objects and motion types. It has 60 videos for training and 30 videos for testing. The Videvo dataset contains 80 videos for training and 20 videos for testing. The training videos are all resized to $300 \times 300$. We mix the DAVIS and Videvo training sets to conduct self-regularization learning as in Section \ref{sec:self-reg}.

\textbf{Metrics.} We evaluate the results in two facets: colorization performance and video temporal consistency. The colorization performance is paramount for colorization task. Without good colorization, the temporal consistency will be less of significance. For example, unsaturated images with few colors could result in better consistency. However, such neutral results cannot meet the requirements of good colorization.
To measure the colorization performance, we adopt PSNR and $L_2$ error in $Lab$ color space. Moreover, we also utilize the colorfulness measurement proposed by Hasler and Suesstrunk \cite{colorfulness}, to roughly evaluate the color diversity of the resulting images produced by different methods. For temporal consistency, we adopt warp error proposed in \cite{btc}. However, warp error is uncorrelated with the video color and is easily affected by the performance of flow estimation module used in the measurement. Therefore, we propose a more suitable Color Distribution Consistency index (CDC) to further measure the temporal consistency, which is specially devised for video colorization task. Specifically, it computes the Jensen-Shannon (JS) divergence  of the color distribution between consecutive frames:
\begin{equation}
CDC_t = \frac{1}{3 \times (N-t)} \sum_{c \in \{r,g,b\}} \sum_{i=1}^{N-t} JS(P_c(I^i), P_c(I^{i+t})),
\end{equation}
where $N$ is the video sequence length and $P_c(I^i)$ is the normalized probability distribution of color image $I^i$ across $c$ channel, which can be calculated from the image histogram. $t$ denotes the time step. A smaller $t$ indicates short-term temporal consistency, while larger $t$ indicates long-term temporal consistency. The JS divergence can measure the similarity between two color probability distributions.
Considering the long-term and short-term temporal consistency together, we propose the following index:
\begin{equation}
CDC = \frac{1}{3}(CDC_1 + CDC_2 + CDC_4).
\end{equation}
It takes $t=1$, $t=2$ and $t=4$ into account, which can appropriately reflect the temporal consistency for color distribution. Too large $t$ will lead to much difference in content between the two frames, causing the color distribution to change rapidly. Moreover, we also conducted a user study for subjective evaluation.

\subsection{Implementation Details}
In our implementation, we adopt CIC \cite{cic} and IDC \cite{tog17} as the image-based colorization backbone $\mathcal{G}$. Note that we do not need to train $\mathcal{G}$. When training, the input interval length $N=10$, while $N$ is set to $17$ when testing. The batch size is 4 and the patch size of input frames is $256 \times 256$. The learning rate is initialized to $5e^{-5}$ and is decayed by half every $10, 000$ iterations. The Adam optimizer \cite{adam} is adopted. We use PyTorch framework and train all models using four GTX 2080Ti GPUs.

\begin{figure*}[htbp]
	\centering
	\includegraphics[width=0.98\linewidth]{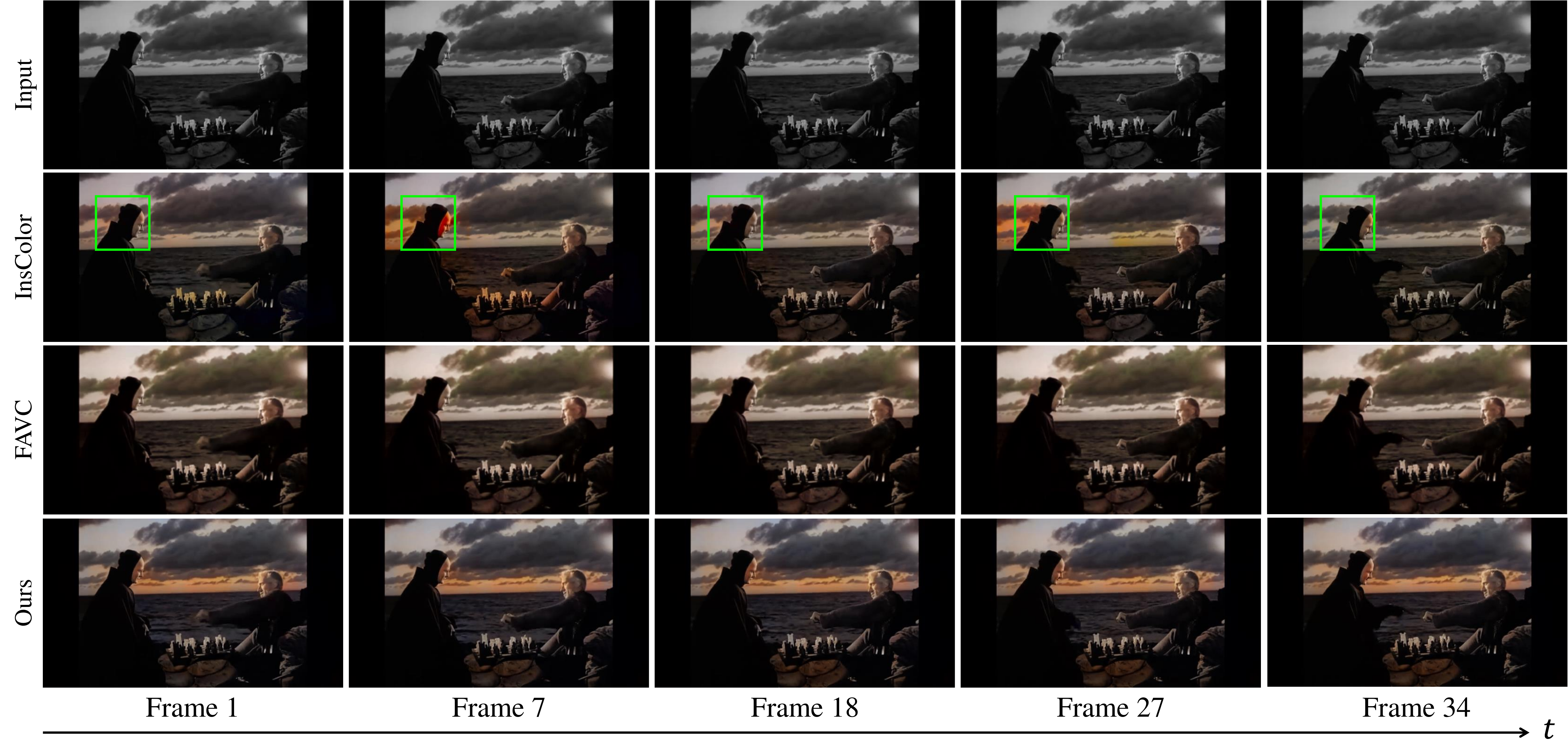}
	\vspace{3pt}
	
	\includegraphics[width=0.98\linewidth]{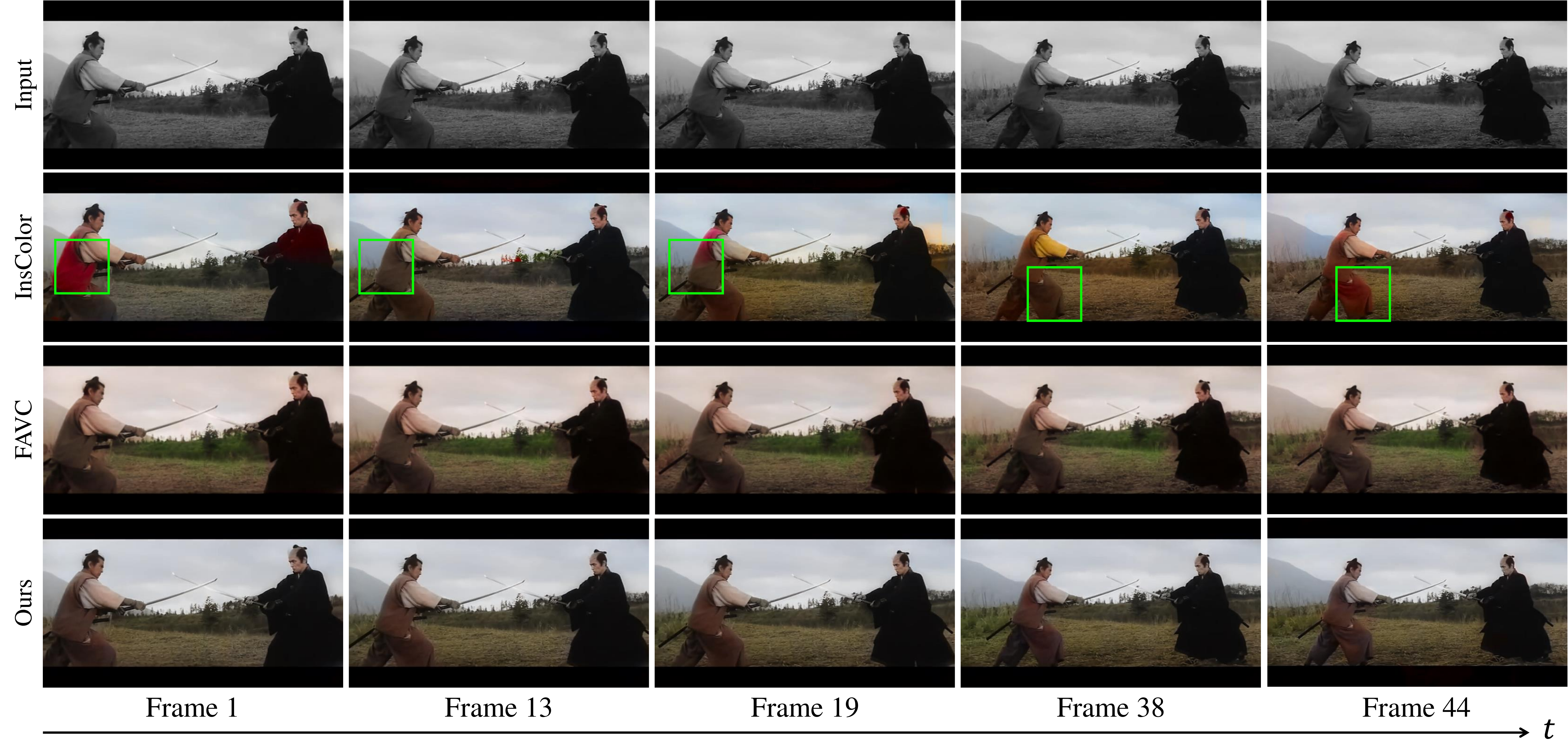}	
	\caption{Visual results on legacy black-and-white movies. Top: \textit{The Seventh Seal (1957)}. Bottom: \textit{Samurai Rebellion (1967)}.}
	\label{fig:real}
	
\end{figure*}

\begin{figure*}[htbp]
	\centering
	\includegraphics[width=0.98\linewidth]{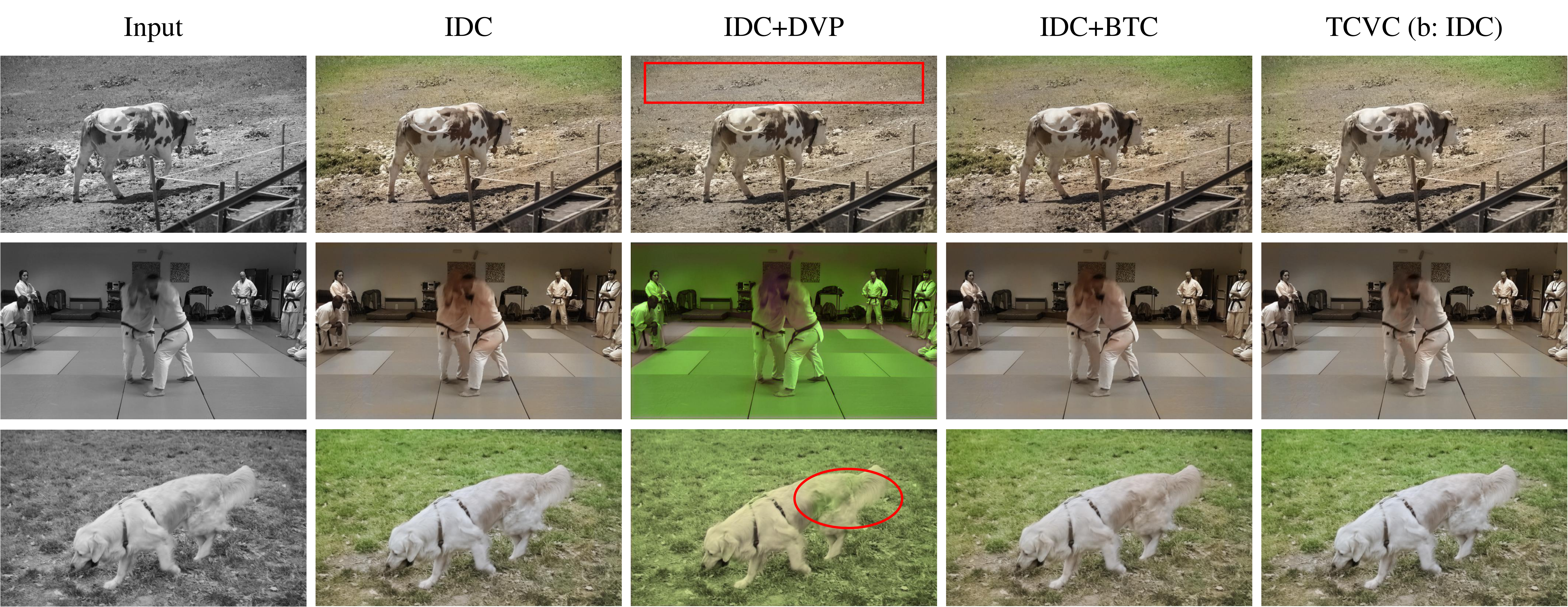}
	\caption{Comparison with post-processing methods BTC \cite{btc} and DVP \cite{dvp}. DVP sometimes could remove the color of the original images (first row) or produce results with weird green tone (second and third rows).}
	\label{fig:compare_DVP}
	
\end{figure*}

\subsection{Comparison with State-of-the-art Methods}
Since this paper focuses on temporally consistent video colorization, FAVC \cite{favc} is the main competitor. FAVC is the newest and the first learning-based fully automatic video colorization method. Unfortunately, there is only FAVC published in the top conference or journal. We follow FAVC and conduct sufficient comparisons with image-based, video-based and post-processing methods. Specifically, we compare our method with representative single image colorization methods \cite{let,cic,tog17,inscolor} and video colorization method FAVC \cite{favc}. In addition, we apply the blind temporal consistency methods BTC \cite{btc} and DVP \cite{dvp} on \cite{cic} and \cite{tog17} to form another two groups of comparison methods. 

\textbf{Quantitative comparison.} The quantitative results are summarized in Table \ref{tab:quantitative_compare} and Figure \ref{fig:chart}. Image-based methods \cite{tog17,inscolor} can achieve relatively higher PSNR, while their temporal consistency is poor. Video-based method FAVC \cite{favc} slightly improves the temporal consistency but its colorization performance is not satisfactory, as shown in Figure \ref{fig:compare}. Quantitatively, FAVC yields the lowest colorfulness value among all the methods. BTC \cite{btc} and DVP \cite{dvp} can largely enhance the temporal consistency, but the cost is that the PSNR values decrease dramatically compared to original outputs of \cite{cic,tog17}. Further, BTC is vulnerable to be affected by outliers and DVP leans to produce colorless results (see Figure \ref{fig:compare_DVP}). Moreover, DVP \cite{dvp} is an image-specific one-shot algorithm which requires independent training during testing. Thus, it is time-consuming to conduct inference, making it impractical for real-time or high-speed applications.

For comparison, we adopt \cite{cic,tog17} as our backbones. After integrated in TCVC, the temporal consistency gets improved, validating the effectiveness of TCVC. Moreover, TCVC can achieve impressive colorization performance with high PSNR values. TCVC can even slightly boost the PSNR values and reduce the $L_2$ error in $Lab$ space. TCVC can also perfectly remain the colorfulness, while BTC and DVP could lower the resulting colorfulness values. Note that, for fairness, we do not use scene cut techniques on the test datasets, but we still achieve the best results. For very long videos, some simple techniques can be used, like histogram/block matching, which are easy to be incorporated with TCVC. With scene cut techniques, the performance of TCVC is supposed to be further improved. 

\textbf{Qualitative comparison.} Visual comparisons are shown in Figure \ref{fig:page_one}, \ref{fig:compare}, \ref{fig:compare1}, \ref{fig:real} and \ref{fig:compare_DVP}. Image-based methods \cite{inscolor, tog17} are prone to produce severe flickering artifacts. Their predicted color of one object differs in consecutive frames. For example, in Figure \ref{fig:compare1}, the car is colorized in red by InsColor \cite{inscolor} in the first four frames , while it is painted bluish in the last frame; The dancer's clothes are colorized in brighter red by IDC \cite{tog17} in the first and fourth frames, while in the other frames, the color of the clothes becomes lighter and less saturated. After applying post-processing method BTC \cite{btc}, the results can become more temporally consistent. However, BTC modifies all the frames of the original output video, which could immensely decrease the PSNR values as discussed before. Further, BTC is susceptible to outliers and cannot deal with the extreme outliers properly and thoroughly. As shown in the lower part of Figure \ref{fig:compare1}, BTC fails to achieve temporal consistency in this consecutive sequence: the outlier red region stays unchanged after applying BTC. As shown in Figure \ref{fig:compare_DVP}, DVP \cite{dvp} could remove the color of the original images or produce results with weird green tone. Further, the results of DVP are likely to contain color contaminations. Compared with state-of-the-art image-based methods, the results of FAVC \cite{favc} are usually not vivid with unsaturated and grayish hue. FAVC \cite{favc} sometimes could even produce strange greenish color in objects (see the lower part of Figure \ref{fig:compare1}). Compared with previous works, our method can achieve both good colorization performance and temporal consistency. Particularly, TCVC can produce colorized results with long-term temporal consistency, since all the internal frames are generated by continual feature propagation. Thus, different from BTC, TCVC can handle outliers and achieve impressive quantitative performance.

\textbf{Results on legacy black-and-white movies.}
Additionally, we display several visual results on legacy black-and-white movies to demonstrates the good generalization ability of our method. It is an attractive application of video colorization. As shown in Figure \ref{fig:real}, one can see that our model is able to produce good colorization results on legacy grayscale films.

\begin{figure}[t]
	\centering
	\includegraphics[width=0.9\linewidth]{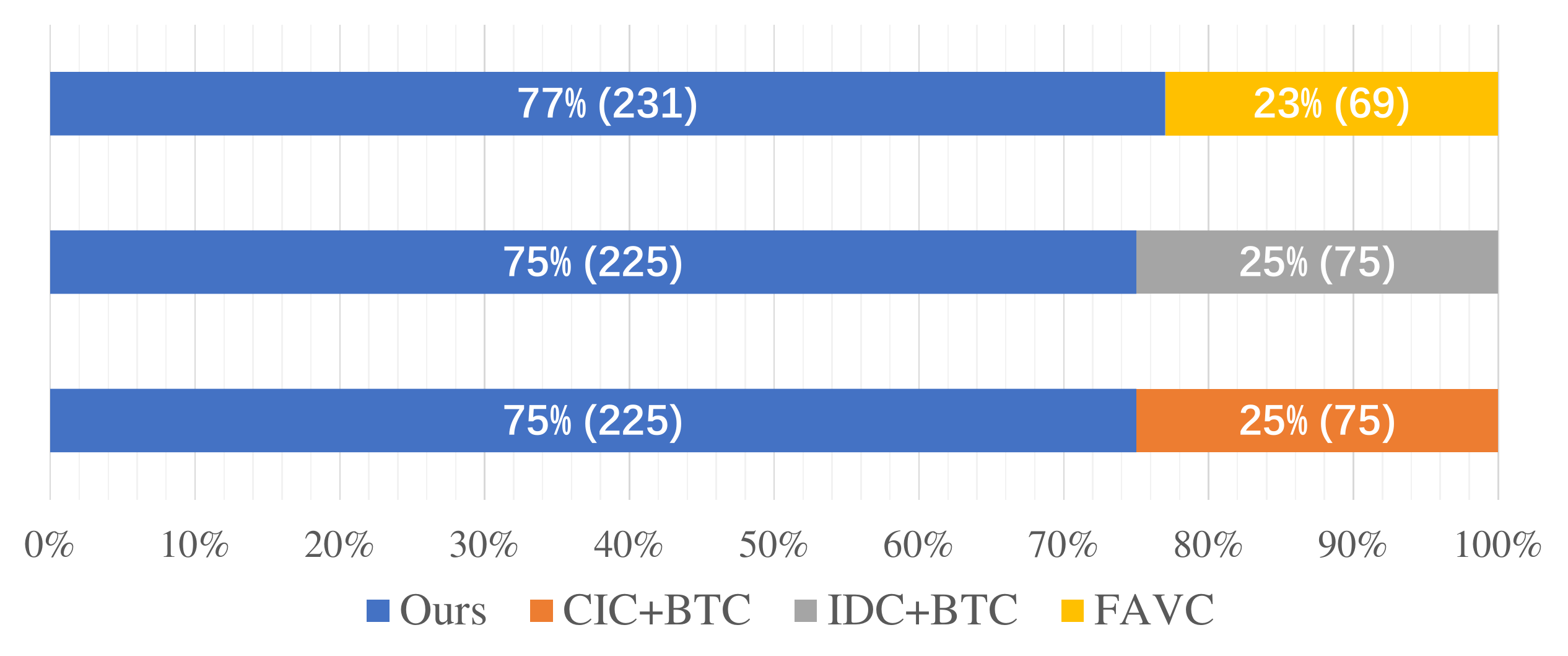}
	\caption{User study results. Users clearly prefer our methods than other state-of-the-art video colorization methods.}
	\label{fig:user_study}
	
\end{figure}

\textbf{User study.}
We also conducted a user study with 20 participates for subjective evaluation. 15 videos are randomly selected from the test datasets. We compare our method with video colorization methods FAVC \cite{favc}, CIC \cite{cic}+BTC and IDC \cite{tog17}+BTC in pairwise manner. The participates are asked to choose the better one by colorization performance and the temporal consistency. As the results shown in Figure \ref{fig:user_study}, the proposed framework TCVC surpasses all other methods by a large margin. More than $75.0\%$ (225) of users' choices favor our results.

\subsection{Advantages of Adopting Anchor Frames}
In TCVC framework, the anchor frames are directly processed by the well-performed image-based model $\mathcal{G}$, and the internal frames are generated by bidirectional propagation from anchor frames. We demonstrate the advantages of adopting anchor frames by statistical analysis. Specifically, we aim to answer the following questions: 1) Since all the anchor frames are the same with that of $\mathcal{G}$, what is the influence of sampling anchor frames with different interval length $N$? 2) What is the effect of adopting deep feature propagation to generate internal frames? 3) What are the advantages of TCVC compared with post-processing method BTC \cite{btc}? To answer these questions, we have calculated the PSNR values of the anchor frames and the internal frames produced by TCVC under different interval length $N$ on DAVIS \cite{DAVIS} dataset. Then, we compare the corresponding PSNR values with the backbone model IDC \cite{tog17} and post-processing method BTC \cite{btc}.

\setlength{\tabcolsep}{6pt}
\begin{table}[h]
	\begin{center}
		\begin{tabular}{c|cccc|c}
			\hline
			$N$ & anchor & \#anchor & internal & \#internal & all \\ \hline
			5 & 25.47 & 372 & 25.40 & 1627 & \multirow{5}{*}{25.42} \\ \cline{1-5}
			9 & 25.43 & 235 & 25.42 & 1764 &  \\ \cline{1-5}
			13 & 25.43 & 184 & 25.42 & 1815 &  \\ \cline{1-5}
			17 & 25.45 & 153 & 25.41 & 1846 &  \\ \cline{1-5}
			21 & 25.47 & 136 & 25.41 & 1863 &  \\ \hline
		\end{tabular}		
	\end{center}
	\caption{Quantitative analysis of image-based colorization backbone IDC \cite{tog17}. We calculate the corresponding PSNR values of anchor frames and internal frames for various $N$. Different $N$ will lead to different separated sets for anchor and internal frames.} \label{tab:IDC}
\end{table}

\setlength{\tabcolsep}{3.8pt}
\begin{table}[h]
	\begin{center}
		\begin{tabular}{c|ccc|l}
			\hline
			& \multicolumn{3}{c|}{PSNR} & \multicolumn{1}{c}{\multirow{2}{*}{CDC}} \\ \cline{2-4}
			& \multicolumn{1}{c|}{anchor} & \multicolumn{1}{c|}{internal} & all & \multicolumn{1}{c}{} \\ \hline
			IDC & 25.45 & 25.41 & 25.41 & 0.005017 \\ \hline
			IDC+BTC & \begin{tabular}[c]{@{}c@{}}24.06\\ (-1.39)\end{tabular} & \begin{tabular}[c]{@{}c@{}}23.90\\ (-1.51)\end{tabular} & \begin{tabular}[c]{@{}c@{}}23.91\\ (-1.5)\end{tabular} & 0.003849 \\ \hline
			TCVC (b: IDC) & \begin{tabular}[c]{@{}c@{}}25.45\\ (0)\end{tabular} & \begin{tabular}[c]{@{}c@{}}25.49\\ (+0.08)\end{tabular} & \begin{tabular}[c]{@{}c@{}}25.49\\ (+0.08)\end{tabular} & 0.003771 \\ \hline
		\end{tabular}		
	\end{center}
	\caption{Effectiveness of TCVC for achieving both good colorization performance and satisfactory temporal consistency. $N=17$.} \label{tab:compare}
	\vspace{-12pt}
\end{table}

Larger interval length $N$ connotes that fewer anchor frames will be sampled. One may concern that TCVC could sample anomalous anchor frames (outliers), resulting in the accumulation of errors throughout the feature propagation. From Table \ref{tab:IDC}, we observe that the PSNR values of anchor frames for different samplings are relatively stable. Statistically, there are few outliers in a sequence. Thus, the probability of sampling anomalous anchor frames is marginal. Further, with the increase of $N$, the number of sampled anchor frames will be reduced, and more outlier frames will fall in internal frames to be regenerated. In such cases, compared with post-processing method like BTC \cite{btc}, TCVC can better get rid of the influence of outliers. As shown in Table \ref{tab:compare}, although BTC could enhance the video temporal consistency, the PSNR is significantly reduced. For TCVC, since the anchor frames are directly processed by image-based method, the PSNR of anchor frames is the same as IDC \cite{tog17}, while the PSNR of internal frames is further improved. This is because TCVC can avoid the influence of anomalous internal frames with low PSNR values, since all the internal frames are regenerated by feature propagation from anchor frames. Hence, TCVC can successfully achieve satisfactory temporal consistency while maintaining good colorization performance.

\section{Ablation Study}
We further conduct ablation studies to demonstrate the effectiveness of the proposed FFM, bidirectional propagation and self-regularization learning. We test the models with interval length $N=11$ on DAVIS dataset.
\subsection{Effectiveness of Feature Fusion Module}
The purpose of the feature fusion module (FFM) is to integrate the backward and forward features in a dedicated corse-to-fine manner. It leverages the information of current frame and adjacent frames to achieve frame-specific feature fusion. To demonstrate its effectiveness, we replace the FFM with plain convolutional networks to fuse the bidirectional features. The experimental results are shown in the second and third rows of Table \ref{tab:component}. By adopting FFM, the temporal consistency is further improved from 0.004003 to 0.003874, which validates the effectiveness of FFM.

\subsection{Effectiveness of Bidirectional Propagation}
In this paper, we propose bidirectional propagation to generate the consecutive internal features from anchor features. If we only conduct unidirectional propagation without complementary information from the opposite direction, the warping operation will cause the features continuously to shift in one direction, leading to information loss. We conduct an ablation study to validate the effectiveness of bidirectional propagation. For unidirectional propagation, we do not need to fuse the forward and backward features with FFM, so we replace FFM with a plain network. As shown in the first and second rows of Table \ref{tab:component}, TCVC model with one direction is much inferior to that with two directions. By the utilization of bidirectional propagation, both the PSNR and temporal consistency are largely improved.

\setlength{\tabcolsep}{10pt}
\begin{table}[h]
	\begin{center}
		\begin{tabular}{lc|cc}
			\hline
			& FFM & PSNR$\uparrow$ & CDC$\downarrow$ \\
			\hline
			one direction & $\times$ &  25.42 & 0.004791\\
			two directions& $\times$ & 25.49  & 0.004003\\
			two directions& $\checkmark$ &  \textbf{25.50} & \textbf{0.003874}\\	
			\hline
		\end{tabular}
	\end{center}
	\caption{Effectiveness of FFM and bidirectional propagation.} \label{tab:component}
\end{table}

\begin{figure*}[htbp]
	\centering
	\includegraphics[width=0.98\linewidth]{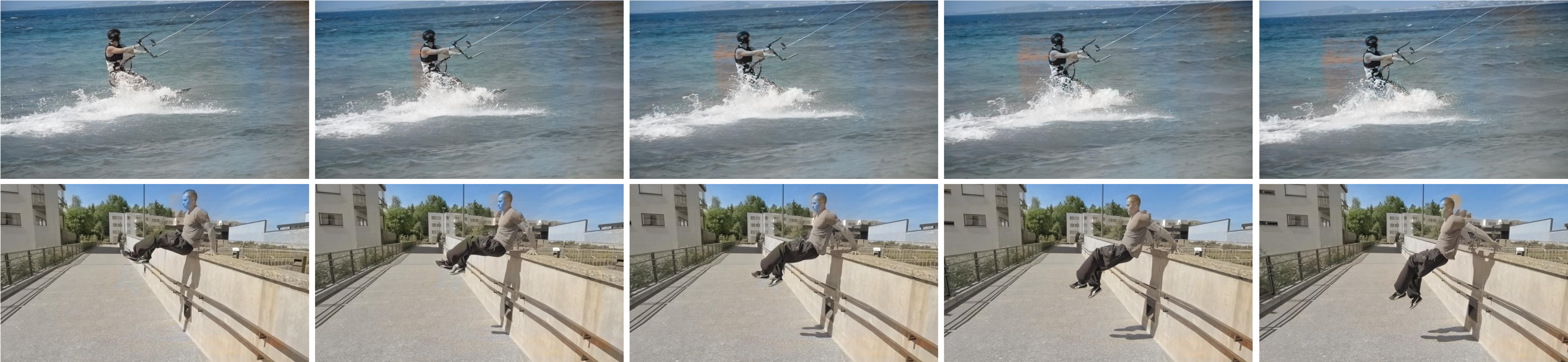}
	\caption{Failure cases of TCVC due to erroneous estimation of optical flow and occulusions.}
	\label{fig:fail}
	
\end{figure*}

\subsection{Effectiveness of Self-regularization Learning}
\begin{figure}[htbp]
	\centering
	\includegraphics[width=0.98\linewidth]{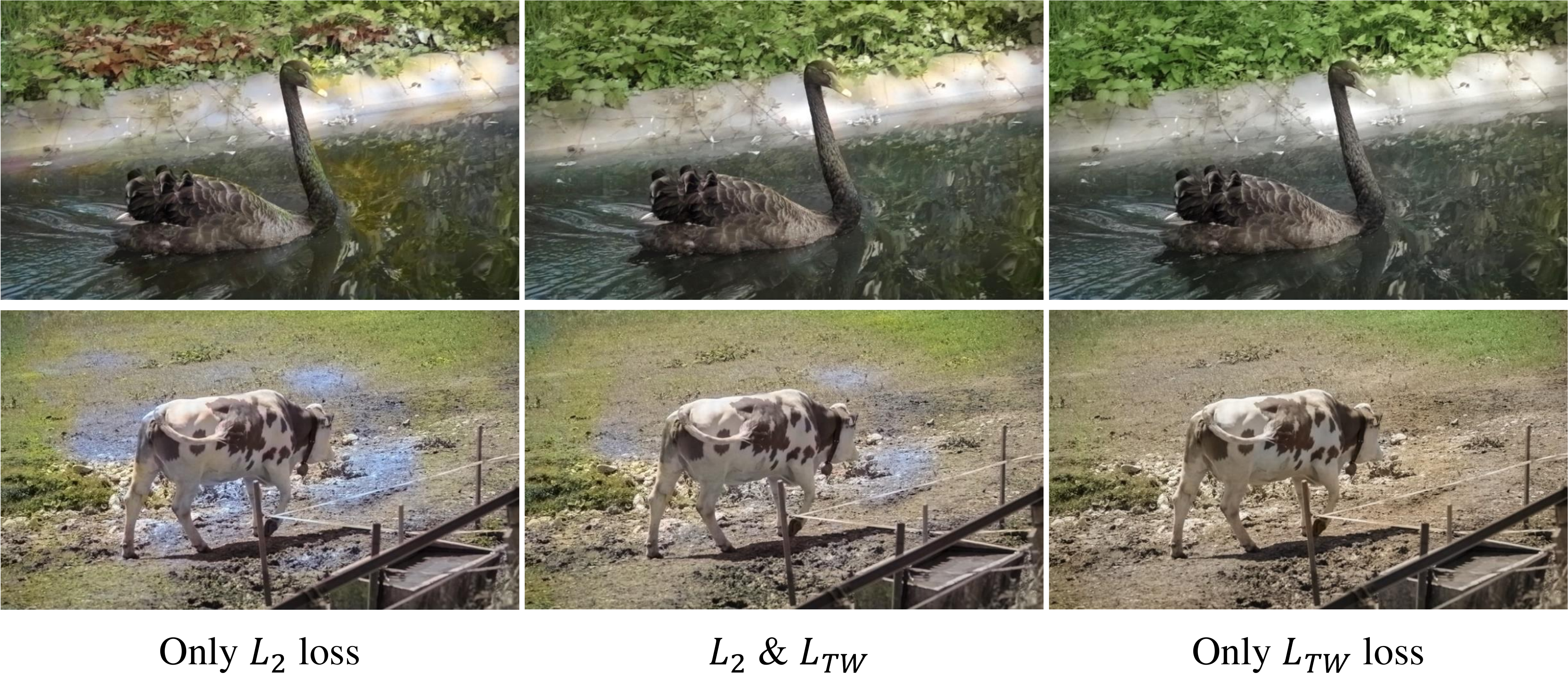}
	\caption{TCVC successfully takes advantage of self-regularization without relying on ground-truth color videos.}
	\label{fig:loss}
	
\end{figure}

We compare conventional supervised-learning and the proposed self-regularization in the same of TCVC framework. In particular, we train the proposed framework using different regularization terms. 1) Only adopting $L_2$ loss with ground-truth color videos. 2) Only adopting temporal warping loss for self-regularization. 3) Adopting both $L_2$ and temporal warping losses simultaneously. As shown in Table \ref{tab:loss}, interestingly, the performance of adopting only $L_2$ loss is much inferior to that of adopting temporal warping loss. It fails to achieve satisfactory colorization performance nor temporal consistency. As shown in Figure \ref{fig:loss}, TCVC model with only adopting $L_2$ loss produces results with severe visual artifacts. This is because $L_2$ loss cannot regularize the procedure of feature warping and fusion in TCVC framework. In addition, $L_2$ loss is not robust to the intrinsic ill-posed nature of colorization problem, which is also addressed in \cite{cic}. When adopting both $L_2$ loss and temporal warping loss, the results are better than only adopting $L_2$ loss but still inferior to only adopting temporal warping loss. Adopting $L_2$ loss on GT will degrade the colorization performance, since we do not retrain $\mathcal{G}$. Avoiding using GT color makes the framework concentrate on reorganizing the consecutive features. This experiment validates the effectiveness of the self-regularization learning for TCVC. It also stresses the difficulties for achieving both good colorization effect and temporal consistency. Therefore, different from conventional supervised learning, we design such an elaborate mechanism, making it efficient and unique.

\setlength{\tabcolsep}{10pt}
\begin{table}[h]
	\begin{center}
		\begin{tabular}{l|cc}
			\hline
			& PSNR$\uparrow$ & CDC$\downarrow$ \\
			\hline
			only $L_{TW}$ &  \textbf{25.50} & \textbf{0.003874}\\
			only $L_2$ &  24.95 & 0.005150\\	
			$L_2$ \& $L_{TW}$ &  25.37 & 0.004536\\
			\hline
		\end{tabular}
	\end{center}
	\caption{Comparison on different learning schemes for TCVC.} \label{tab:loss}
\end{table}

\subsection{Exploration on Interval Length} \label{sec:interval}
To further explore the influence of interval length $N$, we conduct experiments with different $N$ during training and testing phases. In particular, we adopt $N=5$, $N=7$ and $N=10$ for training while $N=3$, $N=5$, $N=9$, $N=11$, $N=17$ and $N=19$ for testing. The experimental results are listed in Table \ref{tab:interval}. It can be seen that the interval length $N$ mainly affects the temporal consistency while it has marginal impact on PSNR values. Adopting more internal frames for training and testing can achieve better temporal consistency. Specifically, when the testing interval length is fixed, the more frames adopted in training, the better the consistency performance. This is because our framework propagates the features in intervals, longer intervals could achieve longer temporal consistency. In addition, adopting more consecutive frames for training can make the model have larger temporal receptive field and learn more motion patterns. Nevertheless, too large interval length $N$ for training will cost more GPU memory and computational resources. Too large $N$ for testing will result in PSNR drop, because the difficulty of optical flow estimation and feature fusion has increased as well. Thus, we adopt moderate $N=10$ for training and $N=17$ for testing in the main experiment.

\setlength{\tabcolsep}{12pt}
\begin{table}[h]
	\begin{center}
		\begin{tabular}{cc|cc}
			\hline
			Train $N$ &Test $N$& PSNR$\uparrow$ & CDC$\downarrow$ \\
			\hline
			\multirow{6}{*}{5} &  3 & 25.47 & 0.004324\\
			&   5 & \textbf{25.49} & 0.004103\\
			&   9 & 25.46 & 0.003949\\
			&  11 & 25.47 & 0.003877\\
			&  17 & 25.48 & \textbf{0.003838}\\
			&  19 & 25.39 & 0.003885\\
			\hline
			\multirow{6}{*}{7} &  3 & 25.47 & 0.004405\\
			&  5 & \textbf{25.50} & 0.004149\\
			&  9 & 25.47 & 0.003951\\
			&  11 & 25.47 & 0.003865\\
			&  17 & 25.46 & 0.003767\\
			&  19 & 25.38 & \textbf{0.003754}\\
			\hline
			\multirow{6}{*}{10} &  3 & 25.47 & 0.004479\\
			&   5 & 25.50 & 0.004227\\
			&   9 & 25.49 & 0.003963\\
			&  11 & \textbf{25.50} & 0.003874\\
			&  17 & 25.49 & 0.003771\\
			&  19 & 25.40 & \textbf{0.003749}\\
			\hline
		\end{tabular}
	\end{center}
	\caption{Exploration on interval length $N$. More internal frames could benefit temporal consistency.} \label{tab:interval}
	\vspace{-10pt}
\end{table}

\section{Failure Cases}
We here show several failure cases of TCVC. As shown in Figure \ref{fig:fail}, in some cases, TCVC could produce results with ghost artifacts or color contamination. This is mainly due to the inaccurate optical flow estimation, especially when large motions or sever occlusions occur. The estimation of optical flow (OF) and occlusion (OCC) is crucial for most video-relevant tasks, e.g., video super-resolution \cite{tof}, video frame interpolation \cite{dain} and video compression \cite{dvc}. However, we have tested TCVC on a large number of videos. It outperforms all other works qualitatively and yields the best quantitative evaluations on average. In the Appendix, we provide the detailed evaluation of each test video. The performance is stable and robust. Certainly, there is room for improving our method. With better OF/OCC estimation, TCVC can continuously be promoted. Research effort on better optical flow and occlusion estimation will largely contribute to lots of computer vision tasks.

\section{Conclusion}
We propose a temporally consistent video colorization framework (TCVC) with deep feature propagation and self-regularization learning. TCVC generates contiguous adjacent features for colorizing video. It adopts a self-regularization learning scheme and does not require any ground-truth color video for training. TCVC can  achieve both good colorization effect and temporal consistency.


%


\ifCLASSOPTIONcaptionsoff
  \newpage
\fi



%
\bibliographystyle{IEEEtran}
\bibliography{IEEEabrv,egbib}

%








\end{document}